\documentclass[10pt,twocolumn,letterpaper]{article}

\usepackage[table,xcdraw]{xcolor}
\usepackage{amsmath}
\usepackage{upgreek}
\usepackage{breqn}
\DeclareMathOperator*{\argmax}{argmax}
\usepackage{booktabs}
\usepackage{subcaption}
\usepackage{wacv}
\usepackage{times}
\usepackage{epsfig}
\usepackage{graphicx}
\usepackage{amsmath}
\usepackage{amssymb}
\usepackage{graphicx}
\usepackage{amsmath}
\usepackage{xcolor}
\usepackage{color, soul}
\usepackage{times}
\usepackage{epsfig}
\usepackage{graphicx}
\usepackage{amsmath}
\usepackage{amssymb}
\usepackage{url}
\usepackage{textcomp}
\usepackage{algorithm}
\usepackage[noend]{algpseudocode}

\usepackage{breakurl}
\makeatletter
\def\BState{\State\hskip-\ALG@thistlm}
\makeatother

\def\wacvPaperID{1081} 

\wacvfinalcopy 

\ifwacvfinal
\fi

\ifwacvfinal
\usepackage[breaklinks=true,bookmarks=false]{hyperref}
\else
\usepackage[pagebackref=true,breaklinks=true,colorlinks,bookmarks=false]{hyperref}
\fi

\ifwacvfinal
\else
\fi

\begin{document}

\title{Action Duration Prediction for Segment-Level Alignment of Weakly-Labeled Videos}

\author{Reza Ghoddoosian \hspace{2cm} Saif Sayed \hspace{2cm} Vassilis Athitsos \\
Vision-Learning-Mining Lab, University of Texas at Arlington\\
{\tt\small \{reza.ghoddoosian, saififtekar.sayed\}@mavs.uta.edu, athitsos@uta.edu
}
}

\maketitle
\ifwacvfinal\thispagestyle{empty}\fi
\begin{abstract}
   This paper focuses on weakly-supervised action alignment, where only the ordered sequence of video-level actions is available for training. We propose a novel Duration Network\footnote{Code available at: https://github.com/rezaghoddoosian/DurNet}, which captures a short temporal window of the video and learns to predict the remaining duration of a given action at any point in time with a level of granularity based on the type of that action. Further, we introduce a Segment-Level Beam Search to obtain the best alignment, that maximizes our posterior probability. Segment-Level Beam Search efficiently aligns actions by considering only a selected set of frames that have more confident predictions. The experimental results show that our alignments for long videos are more robust than existing models. Moreover, the proposed method achieves state of the art results in certain cases on the popular Breakfast and Hollywood Extended datasets.
\end{abstract}

\section{Introduction}

Activity analysis covers a wide range of applications from monitoring systems to smart shopping and entertainment, and it is a topic that has been extensively studied in recent years. While good results have been obtained in recognizing actions in single-action RGB videos ~\cite{trimmed2,trimmed5,trimmed3,trimmed4,2stream,trimmed1}, there are many real-life scenarios where we want to recognize a sequence of multiple actions, whose labels and start/end frames are unknown. Most work done in this area is fully supervised~\cite{fullysupervised3,kuehneWACVend,fullysupervised4,fullysupervised2,RichardStatLang,fullysupervised1,sigurdsson2017asynchronous,fullysupervised5,fullysupervised6}, requiring each frame in the training videos to be annotated. Given the need of deep learning algorithms for ever-larger training datasets,  frame-level annotation can be expensive and unscalable. ``Weak supervision'' is an alternative, where each training video is only annotated with the ordered sequence of actions occurring in that video, with no start/end frame information for any action~\cite{bojanowski,d3tw,TCFPN,ECTC,HMMGMM,hybrid,fine2coarse,NNviterbi}.

\begin{figure}[t]
\begin{center}

   \includegraphics[width=1\linewidth]{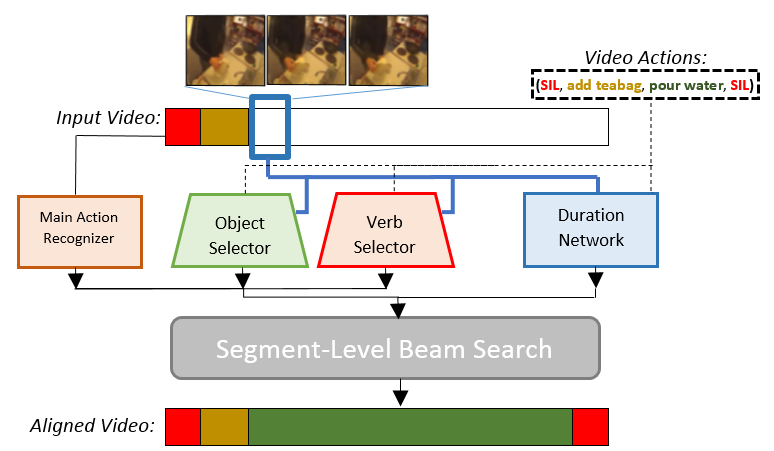}
\end{center}
   \caption{An overview of our proposed method. Based on the context of the temporal window, \textit{pour water} is selected and its duration is predicted to align the given video-level actions }
\label{overview}
\end{figure}

This paper focuses on \textbf{weakly-supervised action alignment}, where it is assumed that the sequence of video-level action labels are provided as input for training and inference, and the output is the start and end time of each action.

A key challenge in weakly supervised action alignment is correctly predicting the duration of actions. To achieve this goal, we propose a \textit{Duration Network (DurNet)} that, unlike previous methods, takes video features into account. Video features contain valuable information that existing duration models ignore. As an example, video features can capture the pace (slow or fast) at which an action is performed. As another example, video features can capture the fact that an ongoing ``frying'' action is likely to continue for a longer time if the cook is currently away from the frying pan. Our duration model learns to estimate the remaining duration of an ongoing action based on the current visual observations. More specifically, the proposed DurNet mainly consists of a bi-directional Long Short-Term Memory (LSTM), which takes as inputs the set of frame features in a short temporal window at a given time, a hypothesized action class and its elapsed duration. The network outputs the probability of various durations (from a discretized set) for the remainder of that action. 

We also introduce a Segment-Level Beam Search algorithm to efficiently maximize our factorized probability model for action alignment. This algorithm modifies the vanilla beam search to predict the most likely sequence of action segments without looping through all possible action-duration combination in all frames. Instead, it predicts the action and duration of segments by selecting a small subset of the frames that are significant enough to maximize the posterior probability.
The time complexity of our Segment-Level Beam Search is linear to the number of action segments in the video, which is theoretically better than that of other Viterbi based alignment methods~\cite{hybrid,RichardStatLang,NNviterbi,CDFL}. In particular Richard \etal~\cite{NNviterbi} considered visual and length models' frame-level outputs and their combinations over all the frames for action alignments. More recently~\cite{CDFL} extended Richard \etal's work~\cite{NNviterbi} by incorporating all invalid action sequences in the loss function during training, but follows the same frame-level inference technique as in~\cite{NNviterbi}.

The main contributions of this paper can be summarized as follows: (1) We introduce a Duration Network for action alignment, that is explicitly designed to exploit information from video features and show its edge over the Poisson model used in previous work~\cite{NNviterbi,CDFL}. (2) We propose a Segment-Level Beam Search that can efficiently align actions to frames without exhaustively evaluating each video frame as a possible start or end frame for an action (in contrast to~\cite{HMMGMM,hybrid,NNviterbi,CDFL}). (3) In our experiments, we use two common benchmark datasets, the Breakfast~\cite{breakfast} and Hollywood Extended~\cite{bojanowski}, and we measure performance using three metrics from~\cite{TCFPN}. Depending on the metric and dataset, our method leads to results that are competitive or superior to the current state-of-the-art for action alignment.

\section{Related Work}
\textbf{Weakly-Supervised Video Understanding.}
Existing methods for video activity understanding often differ in the exact version of the problem that they aim to solve.~\cite{caption2,caption1} aim to associate informative and diverse sentences to different temporal windows for dense video captioning. ~\cite{detection,hide&seek,untrimmed} aim to do action {\em detection}, and are evaluated on videos that consist of typically a single unique action with a large portion of background frames.

Weakly-supervised action segmentation and alignment have been studied under different constraints at training time. Some works utilize natural language narrations of what is happening~\cite{narration2,narration1,whats_cooking,senser,crosstask}. ~\cite{actionset} use only unordered video-level action sets to infer video frames.
Our work is closest to ~\cite{bojanowski,d3tw,TCFPN,ECTC,HMMGMM,hybrid,fine2coarse,NNviterbi,CDFL}, where an ordered video-level sequence of actions is provided for training.

Our paper focuses on the task of weakly-supervised action alignment, where the video and an ordered sequence of action labels are provided as input, and frame-level annotations are the output. 

\textbf{Duration Modeling.} One of the key innovations of our method is in weakly supervised modeling and prediction of action duration. Therefore, it is instructive to review how existing methods model duration. Some methods~\cite{d3tw,TCFPN,ECTC,fine2coarse} do not have an explicit duration model; the duration of an action is obtained as a by-product of the frame-by-frame action labels that the model outputs. ~\cite{jointprediction,whenwhat,Timeconditioned_anticipation} studied long term duration prediction. However they are fully supervised methods whose results are highly sensitive to ground-truth observations.

Most related to our duration model in action alignment are existing methods that model action duration as a Poisson function~\cite{NNviterbi}, or as a regularizer~\cite{bojanowski,length1,hybrid,RichardStatLang} to penalize actions that last too long or too short. Specifically ~\cite{NNviterbi} and ~\cite{CDFL} integrated an action dependent Poisson model into their system which is characterized only by the average duration of each action based on current estimations. The key innovation of our method is that our duration model takes the video data into account. The video itself contains information that can be used to predict the remaining duration of the current action, and our method has the potential to improve prediction accuracy by taking this video information into account.

\section{Method}
In this section, we explain what probabilistic models our method consists of and how they are deployed for our Segment-Level Beam Search.

\subsection{Problem Formulation}
Our method takes two inputs. The first input is a video of $T$ frames, represented by $\mathbf{x}_1^T$, which is the sequence of per-frame features. Feature extraction is a black box, our method is not concerned with how those features have been extracted from each frame. The second input is an ordered sequence $\boldsymbol{\tau} =(\tau_1,\tau_2,...,\tau_M)$ of $M$ action labels, that list the sequence of all actions taking place in the video.

A partitioning of the video into $N$ consecutive segments is specified using a sequence $\mathbf{c}_1^N$ of action labels ($c_n$ specifies the action label for the $n$-th segment) and a sequence $\mathbf{l}_1^N$ of corresponding segment lengths ($l_n$ specifies the number of frames of the $n$-th segment). Given such a partition, we use notation $\pi_n$ for the first frame of the $n$-th segment. 

\begin{figure}[t]
\begin{center}

   \includegraphics[width=1\linewidth]{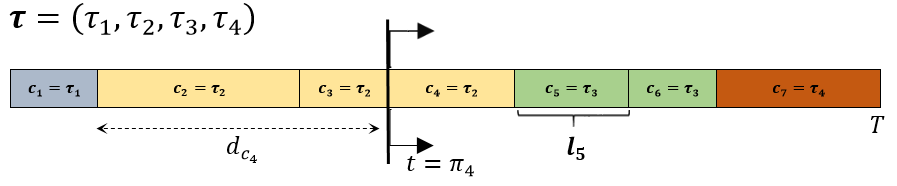}
\end{center}
   \caption{A sample segmented $(N=7)$ video given its video-level labels $\boldsymbol{\tau}$ $(M=4)$. One ground-truth action label $\tau$ can correspond to multiple consecutive segments}
\label{fig_segmentation}
\end{figure}

Given inputs $\mathbf{x}_1^T$ and $\boldsymbol{\tau}$, the goal of our method is to identify the most likely sequence $\overline{\mathbf{c}}_1^N$ of action labels $c_n$ and corresponding sequence $\overline{\mathbf{l}}_1^N$ of durations $l_n$:
\begin{small}
\begin{equation}
\label{equation_argmax}
    (\overline{\mathbf{c}}_1^N,\overline{\mathbf{l}}_1^N) = \argmax_{\mathbf{c}_1^N,\mathbf{l}_1^N} p(\mathbf{c}_1^N, \mathbf{l}_1^N | \mathbf{x}_1^T,\boldsymbol{\tau})    
\end{equation}
\end{small}
We note that $N$ (the number of segments identified by our method) can be different than $M$ (the number of action labels in input $\boldsymbol{\tau}$).
This happens because our method may output the same action label for two or more consecutive segments, and all consecutive identical labels correspond to a single element of $\boldsymbol{\tau}$. We use $\Omega_n$ to denote the earliest segment number such that all segments from segment $\Omega_n$ up to and including segment $n$ have the same action label. For example, in Fig. \ref{fig_segmentation}, $\Omega_4 = \Omega_3 = \Omega_2 = 2$.

Consider a frame $\pi_n$, that is the starting frame of the $n$-th segment. We assume that the remaining duration of an action at frame $\pi_n$ depends on the type of action $c_n$, the elapsed duration $\mathbf{l}_{\Omega_n}^{n-1}$ of $c_n$ up to frame $\pi_n$, and the visual features of a window of $\alpha$ frames starting at frame $\pi_n$. We denote this window as $\mathbf{w}_n = \mathbf{x}_{\pi_n}^{\pi_n+\alpha-1}$. Also, we decompose each action label $c_n$ into a corresponding verb $v_n$ and object $o_n$. For example the action ``take cup'' can be represented by the $(take,cup)$ pair, where $take$ and $cup$ are the verb and object respectively.  
Working with ``verbs'' instead of ``actions'' lets us benefit from the shared information among ``actions'' with the same ``verb''. This specifically helps in analyzing any weakly-labeled video where the frame-level pseudo ground-truth is inaccurate. 
Based on the above, we rewrite $p(\mathbf{c}_1^N, \mathbf{l}_1^N | \mathbf{x}_1^T,\boldsymbol{\tau})$ as:
\begin{small}

\begin{align}
 p(\mathbf{c}_1^N, \mathbf{l}_1^N | \mathbf{x}_1^T,\boldsymbol{\tau})={}& 
 \prod_{n=1}^N p(l_n | \mathbf{w}_n,\mathbf{l}_{\Omega_n}^{n-1},c_n) \cdot p(c_n |  \mathbf{x}_1^{T},\boldsymbol{\tau})\\
={}& \prod_{n=1}^N p(l_n | \mathbf{w}_n,\mathbf{l}_{\Omega_n}^{n-1},v_n,o_n) \cdot p(c_n | \mathbf{x}_1^{T},\boldsymbol{\tau}) \\
={}& \prod_{n=1}^N p(l_n | \mathbf{w}_n,\mathbf{l}_{\Omega_n}^{n-1},v_n) \cdot p(c_n |  \mathbf{x}_1^{T},\boldsymbol{\tau}) \label{equation_terms}
\end{align}
\end{small}
We should note that, in the above equations, in the boundary case where $\Omega_n = n$, we define $\mathbf{l}_{\Omega_n}^{n-1}$ to be 0.
The Duration and Action Selector Network, described next, will be used to compute the probability terms in Eq. \ref{equation_terms}. Then, using our Segment-Level Beam Search, the most likely segment alignment will be identified. 
\begin{figure}[t]
\begin{center}

   \includegraphics[scale=0.5]{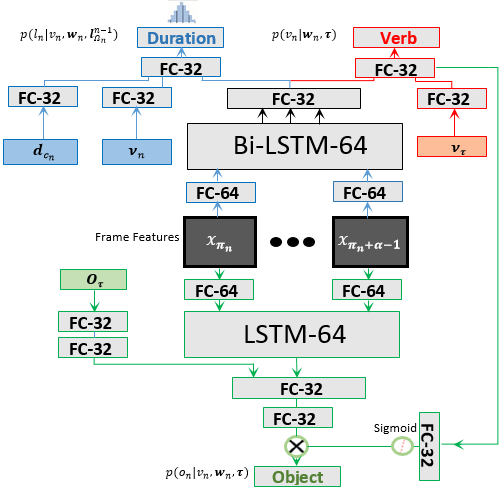}
\end{center}
   \caption{Architecture of the Duration, Object and Verb Selector Networks}\label{architecture} 
\end{figure}

\subsection{Duration Network(DurNet)}\label{section_duration}
Previous work~\cite{hybrid,NNviterbi,CDFL} has tried to model the duration of actions. Richard \etal~\cite{NNviterbi} have used a class-dependent Poisson distribution to model action duration, assuming  that the duration of an action only depends on the type of that action. In contrast, we propose a richer duration model, where the length of an action segment depends not only on the type of that action, but also on the local visual features of the video, as well as on the length of the immediately preceding segments if they had the same action label as the current segment (Eq. \ref{equation_terms}).

The proposed model allows the estimate of the remaining length of an action to change based on video features. For example, our model can potentially predict a longer remaining duration for the action ``squeeze orange'' if the local visual cues correspond to a person just picking up the orange, compared to a person squeezing the orange.

In our method, the range of possible durations of a given action depends on the $verb$ of that action. For example, one second could be half of a short action associated with verb ``take'' and only one-hundredth of a longer action associated with verb ``frying''. We model this dependency by mapping time length to progress units for each verb.  We denote by $\gamma_{v}$ the median length of verb $v$ across all training videos,  and by $L$ the number of time duration bins. We should note that the system cannot know the true value of $\gamma_{v}$, since frame-level annotations are not part of the ground truth. Instead, our system estimates $\gamma_{v}$ based on pseudo-ground truth that is provided using an existing weakly supervised action alignment method, such as \cite{TCFPN,NNviterbi}. Given this estimated $\gamma_{v}$, we discretize the elapsed and remaining time lengths into verb-dependent bins; i.e.\  the bin width ${b}_{v}$ is calculated based on the type of each verb:
\begin{small}
\begin{equation}
    {b}_{v}=\dfrac{\gamma_{v}}{\lfloor\dfrac{L}{2}\rfloor+1}
\end{equation}
\end{small}

The above equation assures that the median length of a verb falls on or around the middle bin, which creates a more balanced distribution for learning.

In our method, $p(l_n | \mathbf{w}_n,\mathbf{l}_{\Omega_n}^{n-1},v_n) $ is modeled by a Bi-LSTM network preceded by a fully-connected layer and followed by fully connected layers and a softmax function $\sigma$ as shown in Fig. \ref{architecture}.
The input to this network, for any segment $n$ at a given time $\pi_n$ , is the one-hot vector representation of the verb $\mathbf{{v}}_{n}\in \mathbb{R}^V$ of a given action $c_n$ and its discretized elapsed duration $\mathbf{{d}}_{c_n}\in \mathbb{R}^L$ as well as the local visual features $\mathbf{w}_n\in \mathbb{R}^{\Gamma \times F}$. Here, $V$ is the total number of verbs, $F$ is the input feature dimension, and $\Gamma$ is the number of temporally sampled features over $\alpha$ frames starting with frame $\pi_n$. At the end, this network outputs the corresponding verb-dependent future progress probability corresponding to each bin. This probability is expressed as an $L$-dimensional vector $\mathbf{k}_{v_n}$, whose $i$-th dimension is the probability that the duration of action $c_n$ falls in the $i$-th progress unit for verb $v_n$, given the inputs described above.

 During training, we used a Gaussian to represent the progress probability labels as soft one-hot vectors. This representation considers the bins that are closer to the true bin more correct than the further ones. The resulting labels are used to compute the standard cross-entropy loss, as the DurNet loss function.
 
Finally, we translate this progress indicator back to time expressed as number of frames, according to verb-dependent steps ${s}_{v}$:
\begin{small}
\begin{equation}
    {s}_{v}=\lfloor\dfrac{\gamma_{v}}{L}\rfloor
\end{equation}
\end{small}
\begin{small}
\begin{equation}
    {l}_{v,i}=(i+1)*{s}_{v}, i \in\{0,1,...,L-1\}
\end{equation}
\end{small}
Thus, the i-th discretized duration $l_{v,i}$ for verb $v$ corresponds to the i-th dimension of vector $\mathbf{k}_{v_n}$, and the value of $\mathbf{k}_{v_n}$ in the i-th dimension gives the probability of discretized duration $l_{v_n,i}$.

\subsection{Action Selector Network}
This network selects the label of the action occurring at any time in the video. Each action is decomposed as a $(verb,object)$ pair. The importance of objects and verbs in action recognition has been studied before~\cite{asynchronous,crosstask}. For example, the verb ``take'' in both ``take bowl'' and ``take cup'' is expected to visually look the same way. These two actions only differ in their corresponding objects. This approach has the advantage that not only the network can access more samples per class (verb/object), but also classification is done over fewer number of classes, because several actions share the same verb/object. This is specifically helpful in weakly-labeled data as the frame-level ground truth is not reliable. The probability of the selected action is obtained by the factorized equation below:
\begin{small}
\begin{equation}
\hspace{0.12cm} p(c_n | \mathbf{x}_1^{T},\boldsymbol{\tau}):=
\eta  [p(o_n | v_n,\mathbf{w}_n,\boldsymbol{\tau})^\zeta p(v_n | \mathbf{w}_n,\boldsymbol{\tau})^\beta p(c_n | \mathbf{x}_1^{T})^\lambda ]
\label{equation_action}
\end{equation}
\end{small}

$\eta[\,]$ is a normalization function that assures :
\begin{small}
\begin{equation}
\sum_{c_n \in \boldsymbol{\tau}} [p(c_n | \mathbf{x}_1^{T},\boldsymbol{\tau})] = 1
\end{equation}
\end{small}
 
 The Action Selector Network consists of three components: i) The verb selector network. ii) The object selector network. iii) The main action recognizer (Fig. \ref{overview}). The influence of each network is adjusted by the $\zeta,\beta $ and $\lambda $ hyper parameters.

 \textbf{i) The Verb Selector Network(VSNet):} It focuses only on the local temporal features during the given time frame [$\pi_n$,  $\pi_n+\alpha-1$] to select the correct verb $v_n$ for segment $n$. The video-level verb labels $\mathbf{v}_{\boldsymbol{\tau}} \in \{0,1\}^V$ are also given as input to the network, where for every $i\in \{0,1,...,V-1\}$, $v_{\boldsymbol{\tau} i}=1$ if $v_{\boldsymbol{\tau} i}$ is present in the video-level verbs, otherwise $v_{\boldsymbol{\tau} i}=0$. 
 
  \textbf{ii) The Object Selector Network(OSNet):} Similar to the VSNet, using the local temporal features, this module selects the correct segment object $o_n$ from the set of video-level objects $\mathbf{o}_{\boldsymbol{\tau}} \in \{0,1\}^O$, where $O$ is the number of available objects in the dataset. Selecting the target object is also influenced by the type of the verb for a given action according to Eq. \ref{equation_action}. In order to model this dependency, latent information from the VSNet flows into the OSNet (Fig. \ref{architecture}).
  
  \textbf{iii) The Main Action Recognizer(MAR):} 
 Unlike the other two components, this module produces frame-level probability distribution for the main actions. This network is more discriminative than the other two and particularly helpful in videos with repetitive verbs and objects. Note that the MAR module can be replaced by any baseline neural network architecture like CNNs or RNNs.

  Finally, as shown in Eq. \ref{equation_action}, the probability of a segment action is defined by fusing the output of the three above-mentioned networks. In the special case of $\zeta,\beta =1$ and $\lambda =0$, the definition of Eq. \ref{equation_action} would be truly probabilistic, and there would be no need for the normalization function $\eta$. The contribution of each network is quantitatively shown in Sec. \ref{action_components_section}. It is noteworthy to mention that our method is equally applicable without the verb-object decomposition assumption. In case there is no specific object associated with actions, our formulation still stands by setting $\zeta =0$ and working with the actions as our set of verbs.

\subsection{Segment-Level Beam Search}

We introduce a beam search algorithm with beam size $B$ to find the most likely sequence of segments, as specified by a sequence of labels $\mathbf{c}_{1}^{N}$ and a sequence of lengths $\mathbf{l}_{1}^{N}$. By combining Eq. \ref{equation_argmax} with Eq. \ref{equation_terms} we obtain:

\begin{small}
\begin{equation} \label{equation_beamsearch}
(\overline{\mathbf{c}}_1^N,\overline{\mathbf{l}}_1^N)=  \argmax_{\mathbf{c}_1^N,\mathbf{l}_1^N} \{\prod_{n=1}^N p(l_n | \mathbf{w}_n,\mathbf{l}_{\Omega_n}^{n-1},v_n) \cdot p(c_n |  \mathbf{x}_1^{T},\boldsymbol{\tau})\} 
\end{equation}
\end{small}

\begin{figure}[t]
\begin{center}

   \includegraphics[width=1 \linewidth]{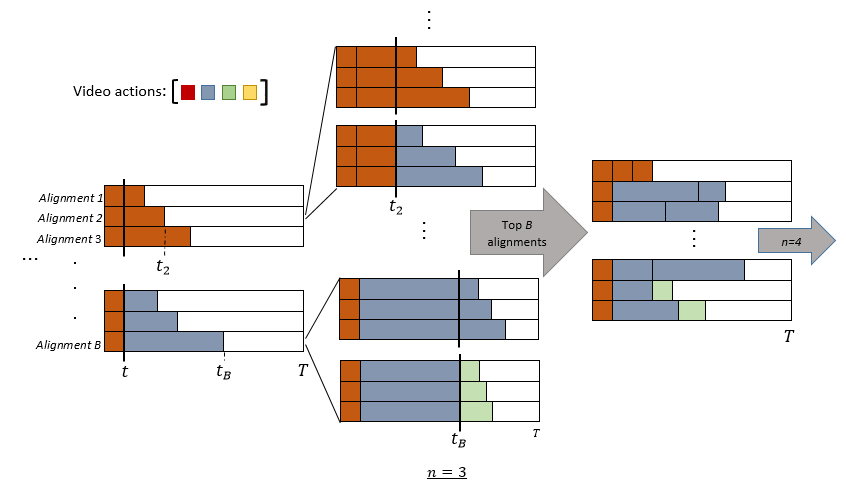}
\end{center}
\caption{Our proposed Segment-Level Beam Search of beam size $B$ during the estimation of the third segment $(n=3)$. For each alignment, different possibilities of next action and its predicted duration are evaluated. At each point in our method, all $B$ hypothesized alignments consist of the same number of segments}
\label{sample}
\end{figure}
In frame-level beam search, different sequences of action classes are considered at every single frame until the end of the video. In contrast, our Segment-Level Beam Search allows the algorithm to consider such sequences only at the beginning of every segment. This technique is inspired by the fact that actions do not change rapidly from one frame to another.

We introduce the notation $A_i(c,l,t_{i})$ to represent the probability of segment-level alignment $i$ until frame $t_i$ for each video, where $c$ and $l$ are the action class and length of the last segment.
We also define $\max^B\{a_1,a_2,...,a_n\}$ as the set of $B$ greatest $a_i$, and calculate $A_i(c,l,t_{i})$ of alignment $i$ recursively for every action $c_n$ and length $l_n$ of segment $n$.
Then, the $B$ most probable alignments with $n$ segments are selected over all combinations of $c_n$ and $l_n$. Algorithm 1 summarizes the procedure for our proposed Segment-Level Beam Search with the following constraints:
\begin{itemize}
  \item $c_1= \tau_1$, $c_N= \tau_M$
  \item $t_i \leq T, \, \forall i\in\{1,2,...,B\}$
\end{itemize}
 $\phi(c_{n-1})$ refers to the set of possible actions  for segment $n$. $\phi(c_{n-1})$ is either a repetition of the action $c_{n-1}$ of the previous segment or the start of the next action in $\boldsymbol{\tau}$. The final segment labels $\mathbf{c}_1^N$ and $\mathbf{l}_1^N$ are derived by keeping track of the maximizing arguments $c_n$ and $l_n$ in the maximization steps.

\begin{algorithm}
\caption{Segment-Level Beam Search}\label{alg:Blink Retrieval Algorithm}
\textbf{Input:}{ Video features $\mathbf{x}_1^T$ and video-level labels $\boldsymbol{\tau}$, beam size B} \\
\textbf{Output:}{ Action label and length sequences $c_1^N$ and $l_1^N$.} 
\begin{algorithmic}
\State $\scriptstyle n \gets 1 $, \Comment{first segment}
\For{$l_1 \in \{l_{v_1,0}, ... ,l_{v_1,L-1}\}$}:
\State ${\scriptstyle A_1(\tau_1,l_1,l_1)=p(l_1 | \mathbf{x}_{1}^{\alpha},0,v_1)\cdot p(\tau_1 | \mathbf{x}_1^{T},\boldsymbol{\tau}) }$ 
\EndFor
\State ${\scriptstyle \mathbb{A}(n)= \max_{l_1}^{B} \{A_1(\tau_1,l_1,l_1)  \}}$, \Comment{set of candidate alignments }\\
\While{$t_i<T$, $\forall i\in \{1,2,...,B \}$}:
\State ${\scriptstyle n \gets n+1}$,
\For{$i\gets $1 to ${B}$}:
\For{ all $c_n \in \phi(c_{n-1}) $;\,$l_n \in \{ l_{v_n,0}$ ... $l_{v_n,L-1}\} $}:

\hspace{\parindent}${\scriptstyle A_i(c_n,l_n,t_i+l_n)=}$ \Comment{$ \scriptstyle A_i(c_{n-1},l_{n-1},t_i)\in \mathbb{A}(n-1)$} \\
\hspace{1.3cm} ${\scriptstyle A_i(c_{n-1},l_{n-1},t_{i})\cdot p(l_n | \mathbf{x}_{t_i}^{t_i+\alpha-1},\mathbf{l}_{\Omega_n}^{n-1},v_n)\cdot p(c_n | \mathbf{x}_1^{T},\boldsymbol{\tau})}$,
\EndFor
\EndFor\\
\hspace{0.5cm}${\scriptstyle \mathbb{A}(n)= \max_{c_n,l_n}^B \{A_i(c_n,l_n,t_i+l_n), \, \forall i\in\{1,2,...,B \} \}  }$
\EndWhile	
\State {$ \scriptstyle A_{\mathrm{final}}(c_N,l_N,T)= \max_{c_N,l_N}^1 \{A_i(c_N,l_N,t_i),\, \forall i\in\{i|t_i=T\} \} $}
\end{algorithmic}
\end{algorithm}

Note that $p(c_n | \mathbf{x}_1^{T},\boldsymbol{\tau})$ in Algorithm 1 is factorized according to Eq. \ref{equation_action}, and every $c_n\in \phi(c_{n-1})$ is broken down to its corresponding $(v_n,o_n)$ pair. This factorization approach encourages segments that cover the whole duration of an action to avoid the penalty each time a new segment is added. This results in faster alignments with a smaller number of unreasonably short segments.

Time complexity of our Segment-Level Beam Search, for each video, depends on the beam size $B$, number of segments $N$ and number of length bins $L$. As $B$ and $L$ are constant values, the time complexity for the algorithm above would be $O(N)$, and only limited to the number of segments per video. Based on our experiments, for the current public action alignment datasets, $N_{max}\approx 70$ is two orders of magnitude less than $T_{max}\approx 9700$. This makes the proposed beam search more efficient than the Viterbi algorithms used in~\cite{NNviterbi,CDFL} and~\cite{hybrid}, which have the complexity of $O(T^2)$ and $O(T)$ respectively.

 \section{Experiments}
We show results on two popular weakly-supervised action alignment datasets based on three different metrics. We compare our method with several existing methods under different initialization schemes. Further, the contribution of each component of our model is quantitatively and qualitatively justified.

\textbf{Datasets.} 1) \textit{The Breakfast Dataset (BD)}~\cite{breakfast} consists of around 1.7k untrimmed instructional videos of few seconds to over ten minutes long. There are 48 action labels demonstrating 10 breakfast recipes with  a  mean  of  4.9  instances  per video. The overall  duration  of the dataset is 66.7h, and the evaluation metrics are conventionally calculated over four splits. 2) \textit{The Hollywood Extended Dataset (HED)}~\cite{bojanowski} has 937 videos of 17 actions with an average of 2.5 non-background action instances per video. There are in total 0.8M frames of Hollywood movies and, following~\cite{bojanowski}, we split the data into 10 splits for evaluation. 

There are four main differences between these two datasets: i) Actions in the \textit{BD} follow an expected scenario and context in each video. However, the relation between consecutive actions in the \textit{HED} can be random. ii) Camera in the \textit{BD} is fixed while there are scene cuts in the \textit{HED}, making the duration prediction more challenging. iii) Background frames are over half of the total frames in the \textit{HED}, while the percentage of them in the \textit{BD} is about 10\%, and iv) The inter-class duration variability in the \textit{BD} is considerably higher than the \textit{HED}.

\textbf{Metrics.}
We use three metrics to evaluate performance: 1) \textit{acc} is the frame-level accuracy averaged over all the videos. 2) \textit{acc-bg} is the frame-level accuracy without the background frames. This is specifically useful for cases where the background frames are dominant as in the \textit{HED}. 3) \textit{IoU} defined as the intersection over union averaged across all videos. This metric is more robust to action label imbalance and is calculated over non-background segments.  

\textbf{Implementation.} For a fair comparison, we obtained the pre-computed 64 dimensional features of previous work~\cite{d3tw,NNviterbi,CDFL}, computed using improved dense trajectories~\cite{iDT} and Fisher vectors~\cite{fisher}, as described in~\cite{kuehneWACVend}.
A single layer bi-directional LSTM with 64 hidden units is shared between the DurNet and VSNet, and a single layer LSTM with 64 hidden units for the OSNet. We followed the same frame sampling as ~\cite{CDFL}, ~\cite{TCFPN} or~\cite{NNviterbi}, depending on the method we use for initialization. We use the cross-entropy loss function for all networks, using Adam optimization~\cite{adam}, learning rate of $10^{-5}$ and batch size of 64. $L$ in the DurNet was set to 7 and 4 for the \textit{BD} and \textit{HED} respectively. In our experiments on the \textit{BD}, we used an $alpha$ of 60 frames and $\zeta$, $\beta$, and $\lambda$ were adjusted to 1, 30, and 5 respectively for our selector network. Beam size in our beam search was set to 150 and other hyperparameters were picked after grid search optimization (refer to supplementary material). 

\textbf{Training Strategy.}
During training, alignment results of a baseline weakly-supervised method, \textit{e.g.} CDFL~\cite{CDFL}, NNViterbi~\cite{NNviterbi} or TCFPN~\cite{TCFPN}, on the training data is used as the initial pseudo-ground truth. We also adopt the pre-trained frame-level action classifier (visual model) of the baseline (CDFL, NNViterbi or TCFPN) as our main action selector component. The initial pseudo-ground truth is used to train our duration and action selector networks. Then, new alignments are generated through the proposed Segment-Level Beam Search algorithm on the training videos. We call these new alignments the ``new pseudo-ground truth''. The adopted visual model is finally retrained based on our ``new pseudo-ground truth'', and used alongside our other components to align the test videos.

\begin{table}
\begin{center}
\footnotesize\setlength{\tabcolsep}{3.2pt}
\caption {\label{comparison} Weakly-supervised action alignment results of existing methods on two main datasets. (* from~\cite{TCFPN}, $^\dag$ best results obtained after running the author's source code multiple times,** after slight changes to the original source code for the specific task.)} 

\begin{tabular}{ l | l l l | l l l}
 & \multicolumn{3}{c}{Breakfast (\%)}  &  \multicolumn{3}{c}{Hollywood Extended (\%)}    \\ 
\cline{2-4}\cline{5-7}  
Models &  acc & acc-bg & IoU & acc &  acc-bg   & IoU  \\    \hline
  \scriptsize{HTK}~\cite{HMMGMM}$^*$  &  43.9 &  \,\,\,\,\, - &  26.6 & 49.4  & \,\,\,\,\, -& 29.1\\
 \scriptsize{ECTC}~\cite{ECTC}$^*$  &  $\sim$35 & \,\,\,\,\, - &\,\,  -  &\,\, - &\,\,\,\,\, -&\,\, - \\
\scriptsize{D$^{3}$TW}~\cite{d3tw}  &  57.0 &\,\,\,\,\,  - &\,\,  -   & 59.4  &\,\,\,\,\, -&\,\, - \\
\hline
 \scriptsize{TCFPN}~\cite{TCFPN}$^\dag$  &  51.7 &\,  48.2 & 33.0  & 57.6  &\, 46.1 & 28.2 \\
\scriptsize{~\cite{TCFPN}/~\cite{NNviterbi} pg}$^{**}$   &  56.4 &\,  53.4 &  36.2 &\,\, -  &\,\,\,\,\, -&\,\, - \\   
\hline
 \scriptsize{NNViterbi}~\cite{NNviterbi}$^\dag$  &   63.5 &\,  63.0 &   \textbf{47.5}  &  59.6  &\, 53.2& 32.4\\
  \scriptsize{~\cite{NNviterbi}/~\cite{TCFPN} pg}$^{**}$   &  63.4 &\,  62.8 &  47.3  &\,\, -  &\,\,\,\,\, -& \,\, - \\ 
\hline
\scriptsize{CDFL}~\cite{CDFL}  &  63.0 & \, 61.4 &  45.8  & \textbf{65.0}$^\dag$  &\, 63.7$^\dag$ & \textbf{40.2}$^\dag$ \\
\hline
\scriptsize{Ours/~\cite{TCFPN} pg}   &  55.7 &\,  56.1 &  36.3 & 50.1  &\, 64.1 & 31.4 \\ 
\scriptsize{Ours/~\cite{NNviterbi} pg}   &  63.7 &\,   65.0 &  42.5  & 56.0  &\,  64.3 &  34.3 \\
\scriptsize{Ours/~\cite{CDFL} pg}  &  \textbf{64.1} & \, \textbf{65.5} &  43.0 & 59.1  & \, \textbf{65.4} & 35.6 \\

\end{tabular}
\end{center}
\end{table}

\subsection{Comparison to State-of-the-Art Methods}

\textbf{Comparison Settings.} In addition to evaluating existing methods, we also evaluate some combinations of existing methods, as follows: 1,2) \textit{Ours/~\cite{NNviterbi} pg} and \textit{Ours/~\cite{CDFL} pg}: Ours initialized with NNviterbi~\cite{NNviterbi} and CDFL~\cite{CDFL} pseudo-ground truth respectively, and a single layer GRU as the MAR. 3) \textit{Ours/~\cite{TCFPN} pg}: Ours initialized with the training results of~\cite{TCFPN} as our pseudo-ground truth, and the TCFPN~\cite{TCFPN} network as the MAR. 4) \textit{~\cite{TCFPN}/~\cite{NNviterbi} pg}: The ISBA+TCFPN method~\cite{TCFPN} initialized with NNViterbi~\cite{NNviterbi} pseudo-ground truth. 5) \textit{~\cite{NNviterbi}/~\cite{TCFPN} pg}: The NNViterbi method~\cite{NNviterbi} initialized with~\cite{TCFPN} pseudo-ground truth.

\textbf{Action Alignment Results.} Table \ref{comparison} shows results for weakly-supervised action alignment. Our method produces better or competitive results for most cases on both datasets. Initialized with CDFL, our method achieves state-of-the-art in two of the three metrics for the Breakfast dataset and in one metric on the Hollywood. We compare our method with CDFL~\cite{CDFL}, NNViterbi~\cite{NNviterbi} and TCFPN~\cite{TCFPN} more extensively, because they are the best open source methods that follow a similar pseudo-ground approach for training. Also for better comparison, in Table \ref{comparison}  we present the results of training NNViterbi on the pseudo ground-truth from TCFPN and vice versa.

In direct head-to-head comparisons with CDFL, NNViterbi and TCFPN, the proposed method often outperforms the respective competitor, and in some cases the head-to-head performance improvement by our method is quite significant. Our method improves action alignment results of TCFPN~\cite{TCFPN} and NNViterbi~\cite{NNviterbi} in 5 (Table \ref{head2head_tcfpn}) and 4 (Table \ref{head2head_nnviterbi}) out of 6 metrics respectively. In addition, we outperform CDFL in frame-level accuracy with and without background on the Breakfast dataset, and when tested on the Hollywood dataset, CDFL accuracy without background is improved while the inference complexity is decreased to $O(N)$ from CDFL's $O(T^2)$ (Table \ref{head2head_cdfl}).

In Table \ref{head2head}, our Segment-Level Beam Search achieves consistent improved results in frame accuracy for both datasets when the background frames are excluded. Considering \textit{acc-bg} is essential especially for the Hollywood dataset as on average around 60\% of the video frames are background, so \textit{acc} values can be misleading. 

There are two plausible explanations on why the performance of our method for non-background actions is not equally repeated for the background segments. First, there is a lack of defined structure in what background can be, which makes it harder to learn. Second, there are cases where background depicts scenes where a person is still or no movement is happening. It is a tough task for even humans to predict how long that motionless scene would last, so the DurNet can easily make confident wrong predictions resulting in inaccurate alignments of background segments. 

\begin{figure}[t]
\begin{center}

   \includegraphics[ width=1\linewidth]{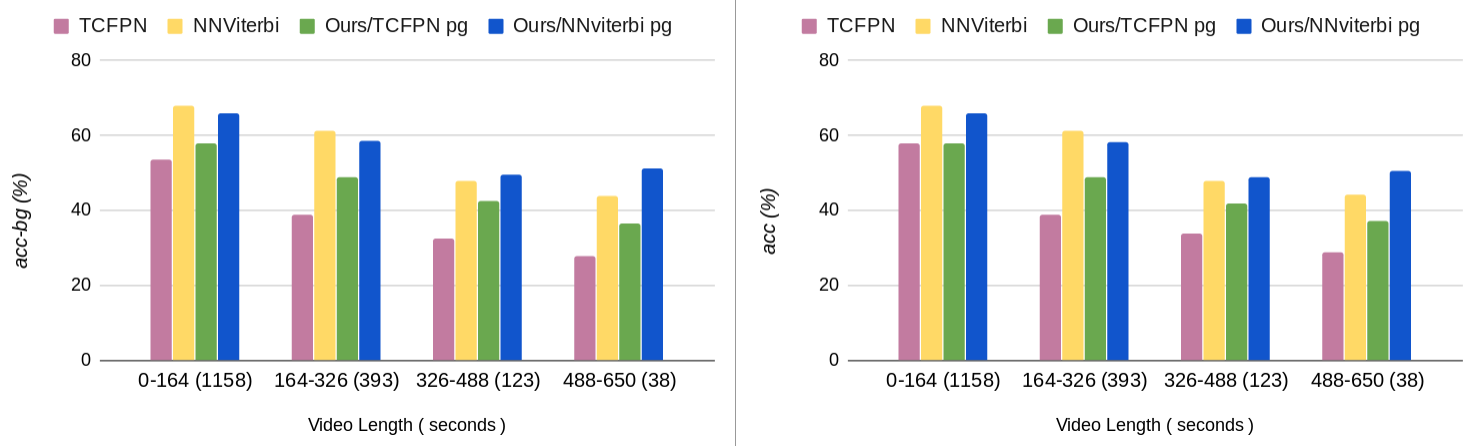}
\end{center}
   \caption{Weakly-supervised action alignment accuracy for videos of different lengths. Unlike the other two baselines, ours is more robust to longer videos. We obtained the results on four equal intervals considering the shortest and longest videos. The number of videos for each interval is mentioned in parentheses} \label{acc_viddur} 
\end{figure}

Fig. \ref{acc_viddur} shows how alignment results vary with video length on the Breakfast Dataset. The performance of our method compared to NNViterbi and TCFPN improves as video length increases. In longer videos, the DurNet can maintain the same action longer depending on the context, while in~\cite{NNviterbi} any duration longer than the action average length gets penalized.

\begin{table}

    \centering
    \footnotesize\setlength{\tabcolsep}{2.3pt}
    \caption{Head-to-head action alignment comparisons of the proposed model with the baselines ($^\dag$  as specified in Table \ref{comparison}).}\label{head2head}
    \subfloat[\label{head2head_tcfpn}]%
      {\begin{tabular}{ l | l l l | l l l}
 & \multicolumn{3}{c}{Breakfast (\%)}  &  \multicolumn{3}{c}{Hollywood Extended (\%)}    \\ 
\cline{2-4}\cline{5-7}  
Models &  acc & acc-bg & IoU & acc &  acc-bg   & IoU   \\    \hline
 \scriptsize{TCFPN}~\cite{TCFPN}$^\dag$  &  51.7 &\,  48.2 & 33.0  & \textbf{57.6}  &\, 46.1 & 28.2 \\
\scriptsize{Ours/~\cite{TCFPN} pg}   &  \textbf{55.7} &\,  \textbf{56.1} & \textbf{36.3} & 50.1  &\, \textbf{64.1} & \textbf{31.4}\\ 
\end{tabular}
      }

    \subfloat[\label{head2head_nnviterbi}]%
      {\begin{tabular}{ l | l l l | l l l}
 & \multicolumn{3}{c}{Breakfast (\%)}  &  \multicolumn{3}{c}{Hollywood Extended (\%)}    \\ 
\cline{2-4}\cline{5-7}  
Models &  acc & acc-bg & IoU  & acc &  acc-bg   & IoU \\    \hline
 \scriptsize{NNViterbi}~\cite{NNviterbi}$^\dag$  &   63.5 &\,  63.0 &   \textbf{47.5}  &  \textbf{59.6}  &\, 53.2& 32.4 \\
\scriptsize{Ours/~\cite{NNviterbi} pg}   &  \textbf{63.7} &\,   \textbf{65.0} &  42.5 & 56.0  &\,  \textbf{64.3} &  \textbf{34.3} \\
\end{tabular}
      }
      
     \subfloat[\label{head2head_cdfl}]%
     {\begin{tabular}{ l | l l l | l l l }
 & \multicolumn{3}{c}{Breakfast (\%)} &  \multicolumn{3}{c}{Hollywood Extended (\%)}    \\ 
\cline{2-4}\cline{5-7} 
Models &  acc & acc-bg & IoU  & acc &  acc-bg   & IoU \\    \hline
\scriptsize{CDFL}~\cite{CDFL}  &  63.0 & \, 61.4 &  \textbf{45.8} & \textbf{65.0}$^\dag$  &\, 63.7$^\dag$ & \textbf{40.2}$^\dag$  \\
\scriptsize{Ours/~\cite{CDFL} pg}  &  \textbf{64.1} & \, \textbf{65.5} &  43.0 & 59.1  &\, \textbf{65.4} & 35.6 \\
\end{tabular}
      }

\end{table}

\subsection{Analysis and Ablation Study}
All analysis and ablation study is done using the TCFPN~\cite{TCFPN} pseudo ground-truth initialization. We also ran our ablation study experiments on the Breakfast dataset mainly, because it consists of videos with many actions and high duration variance, so the impact of learning duration can be measured more effectively.

\subsubsection{DurNet vs. Poisson Duration Model.}
We compare our Duration Network with the Poisson length model used in ~\cite{NNviterbi,CDFL}. To compare the two models, we replaced the DurNet in our Segment-Level Beam Search with the Poisson model in~\cite{NNviterbi,CDFL}, while keeping all other parts of our method unchanged. 

\begin{figure}[t]
\begin{center}

   \includegraphics[ scale=0.55]{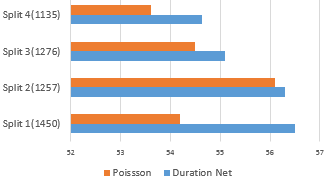}
\end{center}
   \caption{Split-wise frame accuracy on the Breakfast dataset. The number of training videos for each split is indicated in parentheses.}\label{bars} 
\end{figure}

Table \ref{length_comparison} quantitatively shows the advantage of using the context of the video, as it has improved the alignment accuracy by more than 1\%. One reason for the small improvement, however, could be the imbalanced training set size across the four folds. Unlike the statistical Poisson approach, the performance of DurNet, as in other Neural Networks, depends on the training set size. As Figure \ref{bars} shows, the bigger the training data size, the better the performance of the DurNet.

\begin{table}
\begin{center}
\footnotesize\setlength{\tabcolsep}{5pt}
\caption {Comparison between our Duration Network and statistical Poisson length model on the breakfast dataset.}\label{length_comparison} 

\begin{tabular}{ l | l l }
 & \multicolumn{2}{c}{Alignment }   \\ 
\cline{2-3}
Models &  acc & acc-bg  \\    \hline
 \scriptsize{Ours+Poisson}  &  54.56\% &  54.95\%  \\
  \scriptsize{Ours+Duration Net}  &  \textbf{55.70\%} &  \textbf{56.10\%} \\

\end{tabular}
\end{center}
\end{table}

\subsubsection{Duration Step Size Granularity.}
As explained in Section \ref{section_duration}, the predicted durations are discretized into a fixed number $L$ of bins, using different step sizes $s_v$ for different verbs. In order to analyze the advantage of this duration modeling, we compare the weakly-supervised alignment results obtained when we replace this approach with fixed step size for all classes, as well as with different alternatives of adaptive steps (Table \ref{step_comparison}); \ie, the predicted duration range of each action can depend on the maximum, mean or median length of that action calculated across all training videos. A fixed step and a step size dependent on maximum duration, both produce poor results. Step sizes dependent on mean and median durations of actions produce comparable results.

\begin{table}
\begin{center}
\footnotesize\setlength{\tabcolsep}{4.5pt}
\caption { The result of fixed step duration modeling with different alternatives of adaptive steps for weakly-supervised alignment.}\label{step_comparison} 

\begin{tabular}{ l | l  l l}
 & \multicolumn{3}{c}{ Alignment on Breakfast (\%)}  \\ 
\cline{2-4}
Models &  acc & acc-bg   & IoU     \\    \hline
 \scriptsize{Fixed steps($s_v=5$ seconds)}  &  49.9 &\,  49.6 &  32.3   \\
  \scriptsize{Max-based adaptive steps}  & 48.9 &\,  47.6 &  29.7   \\
   \scriptsize{Mean-based adaptive steps}  &  54.9 &\,  55.4 &  35.8   \\
  \scriptsize{Median-based adaptive steps}  &  \textbf{55.7} &\,  \textbf{56.1} & \textbf{36.3}\\

\end{tabular}
\end{center}
\end{table}

\subsubsection{Analysis of the Action Selector Components.}\label{action_components_section}
\begin{figure}[t]
\begin{center}

   \includegraphics[ scale=0.3]{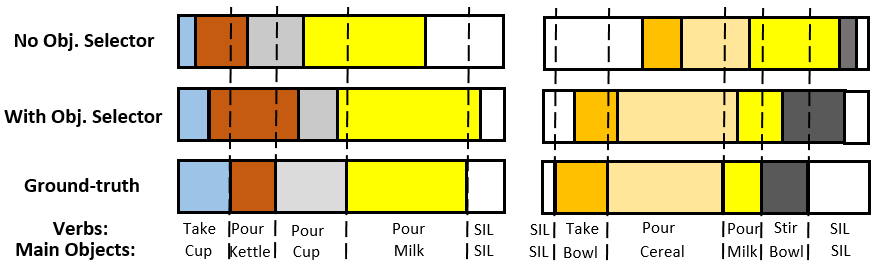}
\end{center}
   \caption{Two sample aligned videos, that consist of action labels with the same verb. The object selector component improves the results by aligning the segments with respect to the correct object. }\label{object_segment} 
\end{figure}

We evaluate the effect of the OSNet, VSNet and MAR separately. Selecting verbs without objects fails in videos where two actions with the same verb happen consecutively in a video, \textit{e.g.}  pour cereal and pour milk (Fig. \ref{object_segment}). Likewise, excluding the VSNet is problematic when two consecutive actions share the same object. Our experiments show that the VSNet and the MAR have the biggest and smallest contributions respectively(Table \ref{component_comparison}).  
We also include the results of the special case where we do not use hyperparameters in Eq. \ref{equation_action}. As we see, a weighted combination of all three components performs best.

\begin{table}
\begin{center}
\footnotesize\setlength{\tabcolsep}{4.5pt}
\caption { Contribution of each action selector component. Having all three components gives the best results.}\label{component_comparison} 

\begin{tabular}{ l | l l l}
 & \multicolumn{3}{c}{ Alignment on Breakfast (\%)}  \\ 
\cline{2-4}
Models &  acc & acc-bg   & IoU    \\    \hline
 \scriptsize{Special case ($\zeta,\beta =1$, $\lambda =0$)}  &  53.9 &\,  54.4 &  35.4  \\
 \scriptsize{Action selector w/o main action}  &  55.5 & \, 56.1 &  36.0 \\
  \scriptsize{Action selector w/o object}  & 54.8 &\,  54.6 &  35.9  \\
   \scriptsize{Action selector w/o verb}  &  50.9 &\,  50.8 &  32.8   \\
  \scriptsize{All components}  &  \textbf{55.7} &\,  \textbf{56.1} & \textbf{36.3} \\

\end{tabular}
\end{center}
\end{table}

\subsubsection{Qualitative Segment-Level Alignment Results.}
One of the benefits of our Beam Search is predicting the class and length of segments without looping through all possible action-length combinations in all frames. Specifically, by predicting the duration of a segment in advance, only a limited set of more significant frames is processed. This leads to faster alignments with competitive accuracy compared to the frame-level Viterbi in~\cite{NNviterbi,CDFL} (Table \ref{head2head}).

\begin{figure}[t]
\begin{center}

   \includegraphics[ width=1.0\linewidth]{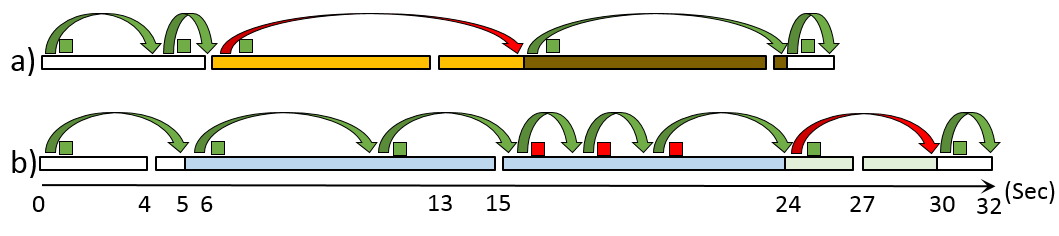}
\end{center}
   \caption{Separate segments denote the ground-truth and the color coded ones indicate the predicted segments. White segments are background. No-background actions are  \textit{add teabag} and \textit{pour water} in video (a), and \textit{pour cereal} and \textit{pour milk} in video (b). }\label{jumps} 

\end{figure}

We demonstrate some success and failure cases of our segment predictions in Fig.\ref{jumps}. It shows how a half minute video can be segmented in a small number of steps. Only a limited window of frames at the start of each step decides the class and length of the corresponding segment. Green and red arrows indicate valid and wrong step duration respectively. Similarly, the correctness of the action selector prediction is shown by the color of the square. 


\begin{figure}[t]
\begin{center}

   \includegraphics[ width=1.0\linewidth]{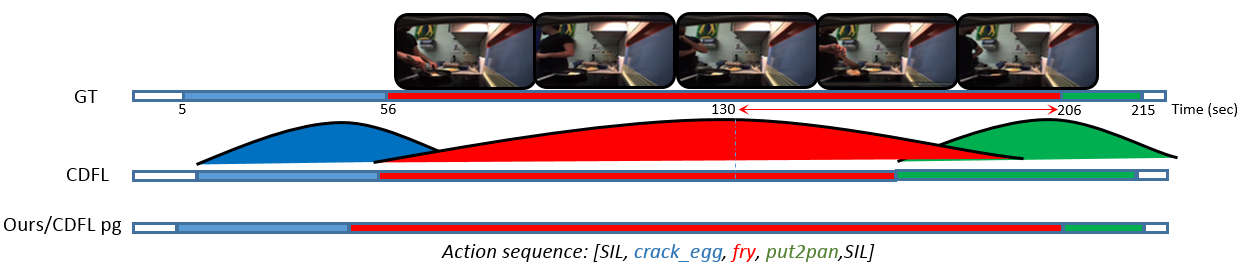}
\end{center}
   \caption{Alignment comparison between CDFL and our method for the test video ``frying egg''. Visual features in DurNet allow the predictions to adapt in duration. The color-coded curves represent the Poisson probability functions characterized by the expected duration of actions in CDFL.} \label{context_length} 
\end{figure}

Finally, Fig.~\ref{context_length} depicts a case where using visual features for length prediction outperforms the Poisson model in~\cite{CDFL}. In this example ``frying'' is done slower than usual due to the subject turning away from the stove and the flipping of the egg. This makes the peak of the Poisson function temporally far from where ``frying'' actually ends resulting in the premature end of the action as longer predictions have very low probabilities and discouraged by the Poisson model. However, our DurNet takes the visual features into account and adapts to longer than expected action durations.

\section{Conclusion}
We have proposed our \textit{Duration Network}, that predicts the remaining duration of an action taking the video frame-based features into account. We also proposed a Segment-Level Beam Search that finds the best alignment given the inputs from the DurNet and action selector module. Our beam search efficiently aligns actions by considering only a selected set of frames with more confident predictions. Our experimental results show that our method can be used to produce efficient action alignment results that are also competitive to state of the art.

\section*{Acknowledgement}
This work is partially supported by National Science Foundation grant IIS-1565328. Any opinions, findings, and conclusions or recommendations expressed in this publication are those of the authors, and do not necessarily reflect the views of the National Science Foundation.

{\small
\bibliographystyle{ieee_fullname}
\bibliography{egbib}

\begin{thebibliography}{10}\itemsep=-1pt

\bibitem{whenwhat}
Yazan Abu~Farha, Alexander Richard, and Juergen Gall.
\newblock When will you do what?-anticipating temporal occurrences of
  activities.
\newblock In {\em Proceedings of the IEEE Conference on Computer Vision and
  Pattern Recognition}, pages 5343--5352, 2018.

\bibitem{narration2}
Jean-Baptiste Alayrac, Piotr Bojanowski, Nishant Agrawal, Josef Sivic, Ivan
  Laptev, and Simon Lacoste-Julien.
\newblock Unsupervised learning from narrated instruction videos.
\newblock In {\em Proceedings of the IEEE Conference on Computer Vision and
  Pattern Recognition}, pages 4575--4583, 2016.

\bibitem{bojanowski}
Piotr Bojanowski, R{\'e}mi Lajugie, Francis Bach, Ivan Laptev, Jean Ponce,
  Cordelia Schmid, and Josef Sivic.
\newblock Weakly supervised action labeling in videos under ordering
  constraints.
\newblock In {\em European Conference on Computer Vision}, pages 628--643.
  Springer, 2014.

\bibitem{length1}
Piotr Bojanowski, R{\'e}mi Lajugie, Edouard Grave, Francis Bach, Ivan Laptev,
  Jean Ponce, and Cordelia Schmid.
\newblock Weakly-supervised alignment of video with text.
\newblock In {\em Proceedings of the IEEE international conference on computer
  vision}, pages 4462--4470, 2015.

\bibitem{trimmed2}
Joao Carreira and Andrew Zisserman.
\newblock Quo vadis, action recognition? a new model and the kinetics dataset.
\newblock In {\em proceedings of the IEEE Conference on Computer Vision and
  Pattern Recognition}, pages 6299--6308, 2017.

\bibitem{d3tw}
Chien-Yi Chang, De-An Huang, Yanan Sui, Li Fei-Fei, and Juan~Carlos Niebles.
\newblock D3tw: Discriminative differentiable dynamic time warping for weakly
  supervised action alignment and segmentation.
\newblock In {\em Proceedings of the IEEE Conference on Computer Vision and
  Pattern Recognition}, pages 3546--3555, 2019.

\bibitem{trimmed5}
C{\'e}sar~Roberto De~Souza, Adrien Gaidon, Eleonora Vig, and Antonio~Manuel
  L{\'o}pez.
\newblock Sympathy for the details: Dense trajectories and hybrid
  classification architectures for action recognition.
\newblock In {\em European Conference on Computer Vision}, pages 697--716.
  Springer, 2016.

\bibitem{TCFPN}
Li Ding and Chenliang Xu.
\newblock Weakly-supervised action segmentation with iterative soft boundary
  assignment.
\newblock In {\em Proceedings of the IEEE Conference on Computer Vision and
  Pattern Recognition}, pages 6508--6516, 2018.

\bibitem{caption2}
Xuguang Duan, Wenbing Huang, Chuang Gan, Jingdong Wang, Wenwu Zhu, and Junzhou
  Huang.
\newblock Weakly supervised dense event captioning in videos.
\newblock In {\em Advances in Neural Information Processing Systems}, pages
  3059--3069, 2018.

\bibitem{trimmed3}
Christoph Feichtenhofer, Axel Pinz, and Richard~P Wildes.
\newblock Temporal residual networks for dynamic scene recognition.
\newblock In {\em Proceedings of the IEEE Conference on Computer Vision and
  Pattern Recognition}, pages 4728--4737, 2017.

\bibitem{trimmed4}
Rohit Girdhar, Deva Ramanan, Abhinav Gupta, Josef Sivic, and Bryan Russell.
\newblock Actionvlad: Learning spatio-temporal aggregation for action
  classification.
\newblock In {\em Proceedings of the IEEE Conference on Computer Vision and
  Pattern Recognition}, pages 971--980, 2017.

\bibitem{ECTC}
De-An Huang, Li Fei-Fei, and Juan~Carlos Niebles.
\newblock Connectionist temporal modeling for weakly supervised action
  labeling.
\newblock In {\em European Conference on Computer Vision}, pages 137--153.
  Springer, 2016.

\bibitem{Timeconditioned_anticipation}
Qiuhong Ke, Mario Fritz, and Bernt Schiele.
\newblock Time-conditioned action anticipation in one shot.
\newblock In {\em Proceedings of the IEEE Conference on Computer Vision and
  Pattern Recognition}, pages 9925--9934, 2019.

\bibitem{adam}
Diederik~P Kingma and Jimmy Ba.
\newblock Adam: A method for stochastic optimization.
\newblock {\em arXiv preprint arXiv:1412.6980}, 2014.

\bibitem{fullysupervised3}
Hilde Kuehne, Ali Arslan, and Thomas Serre.
\newblock The language of actions: Recovering the syntax and semantics of
  goal-directed human activities.
\newblock In {\em Proceedings of the IEEE conference on computer vision and
  pattern recognition}, pages 780--787, 2014.

\bibitem{breakfast}
Hilde Kuehne, Ali Arslan, and Thomas Serre.
\newblock The language of actions: Recovering the syntax and semantics of
  goal-directed human activities.
\newblock In {\em Proceedings of the IEEE conference on computer vision and
  pattern recognition}, pages 780--787, 2014.

\bibitem{kuehneWACVend}
Hilde Kuehne, Juergen Gall, and Thomas Serre.
\newblock An end-to-end generative framework for video segmentation and
  recognition.
\newblock In {\em 2016 IEEE Winter Conference on Applications of Computer
  Vision (WACV)}, pages 1--8. IEEE, 2016.

\bibitem{HMMGMM}
Hilde Kuehne, Alexander Richard, and Juergen Gall.
\newblock Weakly supervised learning of actions from transcripts.
\newblock {\em Computer Vision and Image Understanding}, 163:78--89, 2017.

\bibitem{hybrid}
Hilde Kuehne, Alexander Richard, and Juergen Gall.
\newblock A hybrid rnn-hmm approach for weakly supervised temporal action
  segmentation.
\newblock {\em IEEE transactions on pattern analysis and machine intelligence},
  2018.

\bibitem{narration1}
Ivan Laptev, Marcin Marsza{\l}ek, Cordelia Schmid, and Benjamin Rozenfeld.
\newblock Learning realistic human actions from movies.
\newblock 2008.

\bibitem{fullysupervised4}
Colin Lea, Michael~D Flynn, Rene Vidal, Austin Reiter, and Gregory~D Hager.
\newblock Temporal convolutional networks for action segmentation and
  detection.
\newblock In {\em proceedings of the IEEE Conference on Computer Vision and
  Pattern Recognition}, pages 156--165, 2017.

\bibitem{CDFL}
Jun Li, Peng Lei, and Sinisa Todorovic.
\newblock Weakly supervised energy-based learning for action segmentation.
\newblock In {\em Proceedings of the IEEE International Conference on Computer
  Vision}, pages 6243--6251, 2019.

\bibitem{jointprediction}
Tahmida Mahmud, Mahmudul Hasan, and Amit~K Roy-Chowdhury.
\newblock Joint prediction of activity labels and starting times in untrimmed
  videos.
\newblock In {\em Proceedings of the IEEE International Conference on Computer
  Vision}, pages 5773--5782, 2017.

\bibitem{whats_cooking}
Jonathan Malmaud, Jonathan Huang, Vivek Rathod, Nick Johnston, Andrew
  Rabinovich, and Kevin Murphy.
\newblock What's cookin'? interpreting cooking videos using text, speech and
  vision.
\newblock {\em arXiv preprint arXiv:1503.01558}, 2015.

\bibitem{detection}
Phuc Nguyen, Ting Liu, Gautam Prasad, and Bohyung Han.
\newblock Weakly supervised action localization by sparse temporal pooling
  network.
\newblock In {\em Proceedings of the IEEE Conference on Computer Vision and
  Pattern Recognition}, pages 6752--6761, 2018.

\bibitem{fullysupervised2}
Dan Oneata, Jakob Verbeek, and Cordelia Schmid.
\newblock The lear submission at thumos 2014.
\newblock 2014.

\bibitem{fisher}
Florent Perronnin and Christopher Dance.
\newblock Fisher kernels on visual vocabularies for image categorization.
\newblock In {\em 2007 IEEE conference on computer vision and pattern
  recognition}, pages 1--8. IEEE, 2007.

\bibitem{RichardStatLang}
Alexander Richard and Juergen Gall.
\newblock Temporal action detection using a statistical language model.
\newblock In {\em Proceedings of the IEEE Conference on Computer Vision and
  Pattern Recognition}, pages 3131--3140, 2016.

\bibitem{fine2coarse}
Alexander Richard, Hilde Kuehne, and Juergen Gall.
\newblock Weakly supervised action learning with rnn based fine-to-coarse
  modeling.
\newblock In {\em Proceedings of the IEEE Conference on Computer Vision and
  Pattern Recognition}, pages 754--763, 2017.

\bibitem{actionset}
Alexander Richard, Hilde Kuehne, and Juergen Gall.
\newblock Action sets: Weakly supervised action segmentation without ordering
  constraints.
\newblock In {\em Proceedings of the IEEE Conference on Computer Vision and
  Pattern Recognition}, pages 5987--5996, 2018.

\bibitem{NNviterbi}
Alexander Richard, Hilde Kuehne, Ahsan Iqbal, and Juergen Gall.
\newblock Neuralnetwork-viterbi: A framework for weakly supervised video
  learning.
\newblock In {\em Proceedings of the IEEE Conference on Computer Vision and
  Pattern Recognition}, pages 7386--7395, 2018.

\bibitem{fullysupervised1}
Marcus Rohrbach, Sikandar Amin, Mykhaylo Andriluka, and Bernt Schiele.
\newblock A database for fine grained activity detection of cooking activities.
\newblock In {\em 2012 IEEE Conference on Computer Vision and Pattern
  Recognition}, pages 1194--1201. IEEE, 2012.

\bibitem{senser}
Ozan Sener, Amir~R Zamir, Silvio Savarese, and Ashutosh Saxena.
\newblock Unsupervised semantic parsing of video collections.
\newblock In {\em Proceedings of the IEEE International Conference on Computer
  Vision}, pages 4480--4488, 2015.

\bibitem{caption1}
Zhiqiang Shen, Jianguo Li, Zhou Su, Minjun Li, Yurong Chen, Yu-Gang Jiang, and
  Xiangyang Xue.
\newblock Weakly supervised dense video captioning.
\newblock In {\em Proceedings of the IEEE Conference on Computer Vision and
  Pattern Recognition}, pages 1916--1924, 2017.

\bibitem{sigurdsson2017asynchronous}
Gunnar~A Sigurdsson, Santosh Divvala, Ali Farhadi, and Abhinav Gupta.
\newblock Asynchronous temporal fields for action recognition.
\newblock In {\em Proceedings of the IEEE Conference on Computer Vision and
  Pattern Recognition}, pages 585--594, 2017.

\bibitem{asynchronous}
Gunnar~A Sigurdsson, Santosh Divvala, Ali Farhadi, and Abhinav Gupta.
\newblock Asynchronous temporal fields for action recognition.
\newblock In {\em Proceedings of the IEEE Conference on Computer Vision and
  Pattern Recognition}, pages 585--594, 2017.

\bibitem{2stream}
Karen Simonyan and Andrew Zisserman.
\newblock Two-stream convolutional networks for action recognition in videos.
\newblock In {\em Advances in neural information processing systems}, pages
  568--576, 2014.

\bibitem{fullysupervised5}
Bharat Singh, Tim~K Marks, Michael Jones, Oncel Tuzel, and Ming Shao.
\newblock A multi-stream bi-directional recurrent neural network for
  fine-grained action detection.
\newblock In {\em Proceedings of the IEEE Conference on Computer Vision and
  Pattern Recognition}, pages 1961--1970, 2016.

\bibitem{hide&seek}
Krishna~Kumar Singh and Yong~Jae Lee.
\newblock Hide-and-seek: Forcing a network to be meticulous for
  weakly-supervised object and action localization.
\newblock In {\em 2017 IEEE International Conference on Computer Vision
  (ICCV)}, pages 3544--3553. IEEE, 2017.

\bibitem{fullysupervised6}
Nam~N Vo and Aaron~F Bobick.
\newblock From stochastic grammar to bayes network: Probabilistic parsing of
  complex activity.
\newblock In {\em Proceedings of the IEEE conference on computer vision and
  pattern recognition}, pages 2641--2648, 2014.

\bibitem{iDT}
Heng Wang and Cordelia Schmid.
\newblock Action recognition with improved trajectories.
\newblock In {\em Proceedings of the IEEE international conference on computer
  vision}, pages 3551--3558, 2013.

\bibitem{untrimmed}
Limin Wang, Yuanjun Xiong, Dahua Lin, and Luc Van~Gool.
\newblock Untrimmednets for weakly supervised action recognition and detection.
\newblock In {\em Proceedings of the IEEE conference on Computer Vision and
  Pattern Recognition}, pages 4325--4334, 2017.

\bibitem{trimmed1}
Limin Wang, Yuanjun Xiong, Zhe Wang, Yu Qiao, Dahua Lin, Xiaoou Tang, and Luc
  Van~Gool.
\newblock Temporal segment networks: Towards good practices for deep action
  recognition.
\newblock In {\em European conference on computer vision}, pages 20--36.
  Springer, 2016.

\bibitem{crosstask}
Dimitri Zhukov, Jean-Baptiste Alayrac, Ramazan~Gokberk Cinbis, David Fouhey,
  Ivan Laptev, and Josef Sivic.
\newblock Cross-task weakly supervised learning from instructional videos.
\newblock In {\em Proceedings of the IEEE Conference on Computer Vision and
  Pattern Recognition}, pages 3537--3545, 2019.

\end{thebibliography}
}

\newpage

\section{Implementation Details}
In this section, we provide additional details about our experiments for both Breakfast~\cite{breakfast} and Hollywood Extended~\cite{bojanowski} datasets.

In all our experiments, we trained our three proposed networks (Duration, Verb and Object Selectors) together  with a dropout value of 0.89 and L2 regularization coefficient of 0.0001 for 40 epochs when using~\cite{TCFPN} as our pseudo ground-truth, and 90 epochs when using~\cite{NNviterbi} and~\cite{CDFL} pseudo ground-truth. Our input features were sampled every three frames over $\alpha=60$ frames, at the start of each segment in time.
\subsection{The Breakfast Dataset Experiments}
We set 19 and 14 to be the number of objects and verbs (including background as a separate object/verb) in the Breakfast dataset. $\zeta$, $\beta$, and $\lambda$ were adjusted to 1, 30, and 5 respectively for our selector network using~\cite{NNviterbi} and~\cite{CDFL} as the baseline. In experiments where TCFPN results~\cite{TCFPN} were used as the initial pseudo ground-truth, the aforementioned parameters were slightly changed to 1, 40, and 1.
\subsection{The Hollywood Dataset Experiments}
There are 17 actions (including the background) in the Hollywood Extended dataset, and most of these actions do not share verbs or objects with each other. Hence, it would be redundant to decompose the main actions into their verb and object attributes. As a result, for this dataset, we removed the object selector component and used the 17 main actions as our verbs. $\beta$, and $\lambda$ were set to 3 and 1, and 20 and 1 for the TCFPN~\cite{TCFPN} and NNViterbi~\cite{NNviterbi} baselines respectively. In cases where CDFL~\cite{CDFL} were used, $\beta$ was increased to 50.

Around 60\% of the frames are background in this dataset. Therefore, it is worth mentioning that a naive classifier, that outputs ``background'' for every single frame, can achieve results competitive to the state-of-the-art on the \textit{acc} metric. This is why we emphasize that, specifically for the Hollywood Extended dataset, evaluation using \textit{acc-bg} is more informative. Our method outperforms existing models on this metric while producing better or competitive results on \textit{IoU}.

\subsection{Competitors' Results}
During our observations, we realized that the provided frame-level features are missing for a significant amount of frames in four videos\footnote{1-P34\_cam01\_P34\_friedegg, 2-P51\_webcam01\_P51\_coffee, 3- P52\_stereo01\_P52\_sandwich, 4-P54\_cam01\_P54\_pancake} in the Breakfast dataset. While TCFPN~\cite{TCFPN},  NNViterbi~\cite{NNviterbi} and CDFL~\cite{CDFL} originally trimmed those videos, we decided to remove them for all experiments including our method as well as all baselines~\cite{TCFPN, NNviterbi, CDFL}. 
In Tables 1 and 2 of the main paper, we denote with symbol $\dag$ the best results that we obtained after running the authors' source code for multiple times. The reason we ran the code multiple times is that each training process is randomly initialized and leads to different final result. 

For CDFL~\cite{CDFL} in Table 1 and 2, the alignment \textit{acc-bg} on the Hollywood dataset is somewhat different than the one mentioned in the referenced paper. Similarly, for TCFPN~\cite{TCFPN}, in some cases, our reproduced results are not the same as the ones mentioned in~\cite{TCFPN}. In this case, we reported the results after contacting the authors and having their approval. For a fair comparison in both baselines, we reported the results, that represent the initial pseudo ground-truth in our method. 

Without loss of generality, our final accuracy depends on the quality of the initial pseudo ground-truth, so we have provided the initial pseudo ground-truth and pre-trained main action recognizer models (for TCFPN and NNViterbi on the Breakfast dataset) that we used as supplementary material so our results can be reproduced precisely. All the code and pre-trained models that we provide in supplementary material will be publicly available upon publication of this paper.

\end{document}